\documentclass[10pt,letterpaper,conference]{IEEEtran}
\usepackage{times}
\usepackage{stmaryrd}
\usepackage{amsfonts}
\usepackage{amsmath}
\usepackage{amsthm}
\usepackage{booktabs}
\usepackage{graphicx}
\usepackage{subfigure}
\newtheorem{thm}{Theorem}
\IEEEoverridecommandlockouts

\title{\ \\ \LARGE\bf Analysing Fuzzy Sets Through Combining \\Measures of Similarity and Distance
\thanks{Josie McCulloch, Christian Wagner and Uwe Aickelin are with the School of Computer Science, University of Nottingham, Nottingham, United Kingdom (email: \{psxjm5, christian.wagner, uwe.aickelin\} @nottingham.ac.uk)}
\thanks{This work was partially funded by the EPSRC’s Towards Data-Driven Environmental Policy Design grant, EP/K012479/1 and the RCUK’s Horizon Digital Economy Research Hub grant, EP/G065802/1.}}
\author{Josie McCulloch {\it Student Member, IEEE}, Christian Wagner {\it Senior Member, IEEE} and Uwe Aickelin}
\date{}

\begin{document}

\maketitle

\begin{abstract}
Reasoning with fuzzy sets can be achieved through measures such as similarity and distance. However, these measures can often give misleading results when considered independently, for example giving the same value for two different pairs of fuzzy sets. This is particularly a problem where many fuzzy sets are generated from real data, and while two different measures may be used to automatically compare such fuzzy sets, it is difficult to interpret two different results. This is especially true where a large number of fuzzy sets are being compared as part of a reasoning system. This paper introduces a method for combining the results of multiple measures into a single measure for the purpose of analysing and comparing fuzzy sets. The combined measure alleviates ambiguous results and aids in the automatic comparison of fuzzy sets. The properties of the combined measure are given, and demonstrations are presented with discussions on the advantages over using a single measure.
\end{abstract}

%\begin{IEEEkeywords}
% similarity measure, distance measure, directional distance, fuzzy sets, ordered weighted average (OWA)
%\end{IEEEkeywords}

\section{Introduction}
\label{sec:introduction}
\PARstart{T}{o} compare two fuzzy sets (FSs) one may consider their similarity or distance. To assess their similarity, we measure the similarity of the membership values for each element in each set. The result is given within the interval $[0,1]$, where 0 indicates that there are no elements shared between both FSs and 1 indicates that the sets are identical. Alternatively, to assess the distance between FSs, given as a value in $\mathbb{R}$, we measure the distance between the elements which belong to each set; typically the distance between elements is also weighted by their membership values. 

Measures of similarity and distance have frequently been applied to a variety of different applications. For example, similarity has often been used to measure the similarity between different word models \cite{WuComparative, wagnerwordmodels2013}, or to find similar patterns in classification \cite{Wang20052063} and clustering \cite{Wang20052063}.
Distance Measures (DMs), though less commonly researched, have been used to compare FSs, for example, in the ranking of fuzzy numbers \cite{Cheng1998307}.

Measures of similarity and distance evaluate two fundamentally different aspects of FSs, and it is due to the unique properties of these measures, or more directly, through the nature of \emph{what} the measures actually measure that their applicability to a given problem setting is determined. For example, there are cases in which a similarity measure (SM) may not be useful, such as when the FSs are disjoint. In this case, the result of the SM is always zero. This does not tell us how far apart the FSs are placed in the universe of discourse (UoD); they may be far apart or right next to each other. Where this is of concern, a DM may be beneficial. However, likewise, a DM is also not always a useful measure, for example when one FS is a subset of another. In this case the results become ambiguous as DMs are not ideal for detecting overlap between FSs. 

Current research within the literature has generally made a choice between using either measures of similarity or distance, however in many cases, it is not trivial to make this choice, in particular when FSs are dynamically created from data such as for approaches like \cite{wagnerwordmodels2013} and \cite{coupland2010intervalapproach}. This paper proposes the fusion of both measures into a single measure which can be applied in the comparison of FSs and produces meaningful results regardless of the exact nature of the FSs to be measured. The fusion is achieved by an ordered weighted average (OWA) operator, and is applied to data-driven FSs to demonstrate the benefits of the measure.

Section \ref{sec:background} introduces FSs, SMs, DMs, and OWA operators, followed by an examination of what exactly the measures measure in Section \ref{sec:comparing_measures}. In Section \ref{sec:combining_measures}, a combined measure is presented which utilises the unique properties of both similarity and distance, and demonstrations of the combined measure are shown in Section \ref{sec:demonstrations}. Finally, conclusions are given in Section \ref{sec:conclusions}.

%%%%%%%%%%%%%%%%%%%%%%%%%%%%%%%%%%%%%%%%%%%%%%%%%%%%%%%%%%%%%%%%%%%%%%%%%%%%%%%%%%%%%%%%%%%%%%%

\section{Background}
\label{sec:background}
\subsection{Fuzzy Sets}
\label{sec:fuzzy_sets}
Fuzzy sets have been applied to many applications in which uncertainty is present; some examples of which include data mining \cite{datamining} and Computing with Words \cite{CWW}. Unlike traditional logic, for which the membership of each element to the set is a Boolean value, the elements of a FS have a membership value that lies anywhere in the interval $[0,1]$. A FS $F$ may be represented as a set of ordered pairs as follows \cite{mendel2001uncertain}:
\begin{equation}
 F = (x, \mu_F(x))\ |\ x \in X
\end{equation}
where $\mu_F(x)$ indicates the membership value of the element $x$ in the FS $F$. For a discrete UoD, the FS $F$ may be written as \cite{mendel2001uncertain}
\begin{equation}
 F = \sum_x \mu_F(x)\ /\ x
\end{equation}
where $\sum$ denotes the collection of all points $x \in X$ with associated membership value $\mu_F(x)$.

\subsection{Similarity Measures}
\label{sec:similarity_intro}
SMs are a common tool used within fuzzy logic. A SM $s(A,B) \rightarrow [0,1]$ calculates how similar two FSs are to each other through a comparison of the degrees of membership within each set. Common properties of a SM $s$ for FSs $A$, $B$ and $C$ are as follows:\\
\textbf{Reflexivity:} $s(A, B) = 1 \Longleftrightarrow A = B $ \\
\textbf{Symmetry:} $s(A, B) = s(B, A)$ \\
\textbf{Transitivity:} If $A \subseteq B \subseteq C$, then $s(A, B) \geq s(A, C)$ \\
\textbf{Overlapping:} If $A \cap B \neq \emptyset$, then $s(A, B) > 0$; \\
 otherwise, $s(A, B) = 0$ 
 
 Note that it is not necessary for a SM to have all of these properties as the application for which the measure is used may not depend on all of them. However, it is typical that a SM always follows the property of reflexivity.

Throughout this paper, similarity is measured using the Jaccard SM, which supports all of the four properties listed above \cite{WuComparative}. The Jaccard measure $s$ for FSs $A$ and $B$ is given as:
\begin{equation}
 s(A, B) = \frac{\sum^n_{i=1} min(\mu_A(x_i),\ \mu_B(x_i))}
		{\sum^n_{i=1} max(\mu_A(x_i),\ \mu_B(x_i))}
  \label{eq:jaccard}
\end{equation}
where $n$ is the total number of discretisations along the $x$-axis.

%%%%%%%%%%%%%%%%%%%%%%%%%%%%%%%%%%%%%%%%%%%%%%%%%%%%%%%%%%%%%%%%%%%%%%%%%%%%%%%%%%%%%%%%%%%%%%%
\subsection{Distance Measures}
\label{sec:distance_intro}
A DM $d(A,B) \rightarrow \mathbb{R}^+$ is used to asses the distance between FSs by calculating the distances between the elements in each set.

A DM $d$ on FSs $A$, $B$ and $C$ holds the following properties: \\
\textbf{Self-identity}: $ d(A, A) = 0 $ \\
\textbf{Separability}: $ d(A, B) \geq 0 $ \\
\textbf{Symmetry:} $ d(A, B) = d(B, A)$\\
\textbf{Transitivity:} If $A \subseteq B \subseteq C$, then $d(A, B) \leq d(A, C)$ \\
\textbf{Triangle inequality}: $ d(A, C) \leq d(A, B) + d(B, C) $ 
 
The distance between two FSs is most commonly measured by taking $\alpha$-cuts of FSs and measuring the distance between the $\alpha$-cuts. The $\alpha$-cut of the FS $A$ is a non-FS comprised of all the elements whose membership grade within $A$ is greater than or equal to $\alpha$ \cite{Zadeh1975199}; this is written formally as $A_\alpha = \{x\ |\ \mu_A(x) \geq \alpha \}$.

Chaudhuri and Rosenfeld \cite{Chaudhur19961157} proposed the following metric to measure the distance between two convex, normal FSs $A$ and $B$:
\begin{equation}
d(A, B) = \frac{\sum^m_{i=1} y_{\alpha_i}\ h(A_{\alpha_i}, B_{\alpha_i})}{\sum^m_{i=1} y_{\alpha_i}}
\label{eq:CR_haus}
\end{equation}
where the $y$-axis is discretised into $m$ points ($y_1, y_2, ..., y_m$), $A_{\alpha_i}$ is the non-fuzzy $\alpha$-cut (given as an interval) of the FS $A$ at y-coordinate $y_{\alpha_i}$, and $h$ is the conventional Hausdorff metric for two continuous intervals $\bar{A}$ and $\bar{B}$ as follows \cite{Zwick1987221}:
\begin{equation}
 h(\bar{A}, \bar{B}) = max \{|\bar{A}_l - \bar{B}_l |, |\bar{A}_r - \bar{B}_r |\}
 \label{eq:interval_haus}
\end{equation}
where $\bar{A} = [\bar{A}_l, \bar{A}_r]$ and $\bar{B} = [\bar{B}_l, \bar{B}_r]$.

In addition to the Hausdorff distance given above, a directional DM (DDM) is given as follows \cite{mcculloch2013measuring}:
\begin{equation}
  h(\bar{A}, \bar{B})=
  \begin{cases}
    \bar{B}_l - \bar{A}_l, & \text{if $|\bar{B}_l - \bar{A}_l| > |\bar{B}_r - \bar{A}_r| $}.\\
    \bar{B}_r - \bar{A}_r, & \text{otherwise}.
  \end{cases}
  \label{eq:interval_haus_with_sign}
\end{equation}
for which a positive distance is given where $A < B$, and a negative value of distance is given where $B > A$. The DDM, however, does not hold the property of symmetry and instead follows partial symmetry, defined as $|d(A,B)| = |d(B,A)|$ and $d(A,B) \neq d(B,A)$ where $A \neq B$. Throughout this paper, (\ref{eq:CR_haus}) is used in conjunction with (\ref{eq:interval_haus_with_sign}).

Having reviewed SMs and DMs, a brief overview of OWA operators is given next, which will be used to aggregate similarity and distance.

%%%%%%%%%%%%%%%%%%%%%%%%%%%%%%%%%%%%%%%%%%%%%%%%%%%%%%%%%%%%%%%%%%%%%%%%%%%%%%%%%%%%%%%%%%%%%%%%%%%%%%%

\subsection{Ordered Weighted Average}
\label{sec:owas}
OWA operators \cite{yager1988ordered} are used to aggregate sub-components of a problem. An OWA involves assigning objects to an ordered set of weights $w = \{w_1, w_2, ....., w_n\}$, for which $w_i \in [0, 1]$ and $\sum^n_{i=1} w_i = 1$. The objects which are to be aggregated are sorted into descending order, and each object is multiplied by the corresponding weight. Thus, for a given list of objects $a_1, a_2, ..., a_n$ and weights $w_1, w_2, ..., w_n$, the OWA is calculated as follows \cite{yager1988ordered}:

\begin{equation}
 F(a_1, a_2, .... a_n) = w_1 b_1 + w_2 b_2 + .... + w_n b_n
 \label{eq:OWA}
\end{equation}
where $b_i$ is the $i_{th}$ largest element in the collection $a_1, a_2, ..., a_n$.

OWAs have been commonly used in the literature to solve a variety of problems. For example, \cite{Canos2008669} uses an OWA in decision making applied to the personnel selection problem. In \cite{Sadiq20104881}, an OWA is used to aggregate different performance indicators to assess the performance of small drinking water utilities, and \cite{Sadiq20104881} uses and OWA to aid in the selection of financial products.

%%%%%%%%%%%%%%%%%%%%%%%%%%%%%%%%%%%%%%%%%%%%%%%%%%%%%%%%%%%%%%%%%%%%%%%%%%%%%%%%%%%%%%%%%%%%%%%%%%%%%%%

\section{Comparison of Measures on Fuzzy Sets}
\label{sec:comparing_measures}
In this Section, SMs and DMs are compared on a series of real-data driven FSs. This is in order to clarify their respective outputs in an applied context, and to demonstrate the proposition that it can be more beneficial  to use a combination of both measures. 

\begin{figure}[t!]
 \centering
  \subfigure[]
  {
      \includegraphics[height=4.4cm, width=6cm]{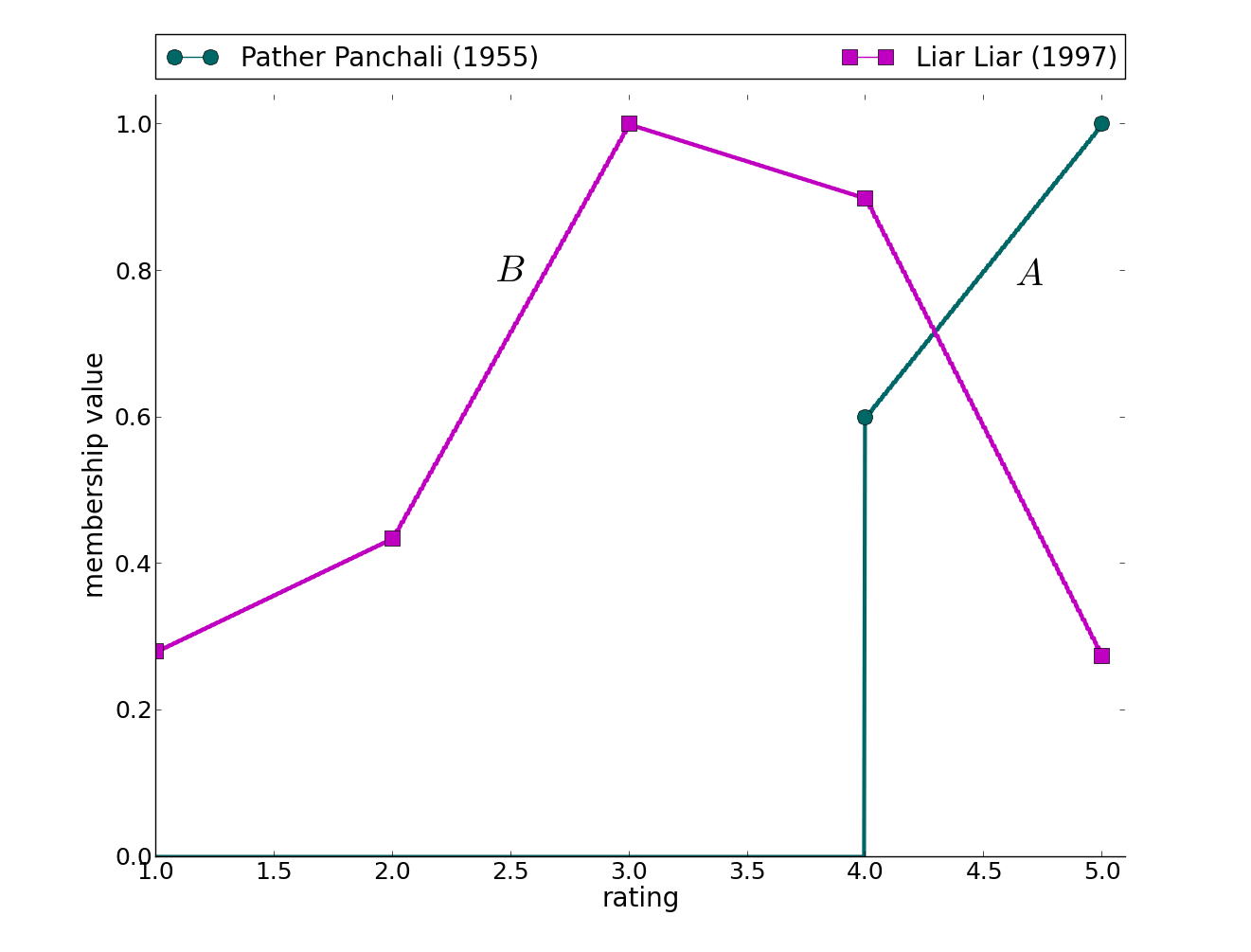}
      \label{fig:demo_d}
  }
  \subfigure[]
  {
      \includegraphics[height=4.4cm, width=6cm]{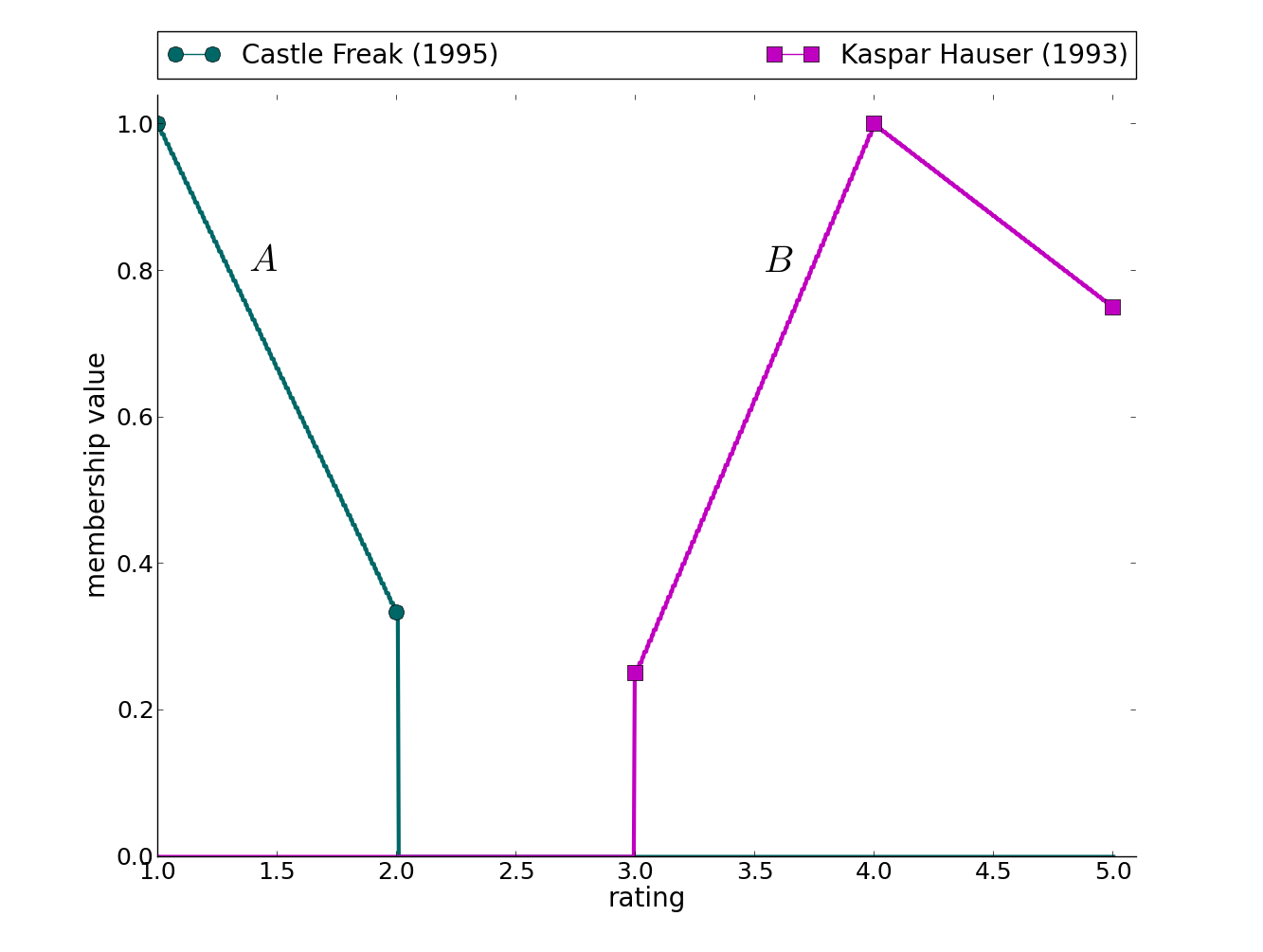}
      \label{fig:demo_e}
  }
  \subfigure[]
  {
      \includegraphics[height=4.4cm, width=6cm]{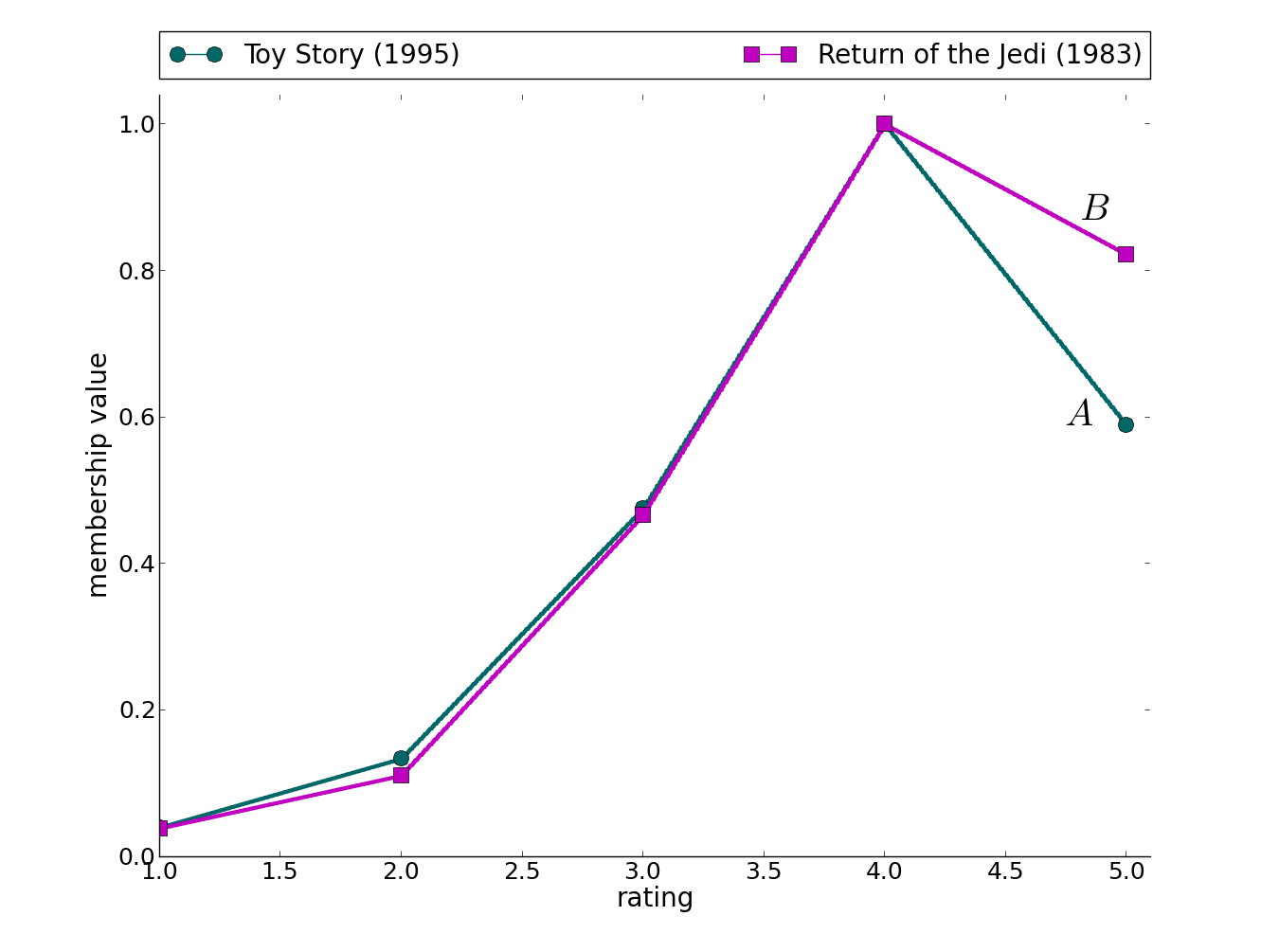}
      \label{fig:demo_f}
  }
  \caption{Fuzzy sets used to demonstrate the attributes of SMs and DMs in Table \ref{tab:type1_results}.}
  \label{fig:demonstration_T1_sets}
\end{figure}

As previously discussed, SMs and DMs have unique properties which lead to them measuring fundamentally different concepts. To demonstrate the nature of the measures, and the strengths of using both similarity and distance together to analyse FSs, consider the FSs shown in Fig. \ref{fig:demonstration_T1_sets}. These FSs have been constructed from the Movie Lens data set \cite{movielens}, in which films are rated between 1 (poor) and 5 (great). Histograms were created to represent the distribution of ratings and each histogram was normalised by dividing the membership value at each $x$-coordinate by the peak membership value of the histogram. Linear-interpolation was used to determine membership values between known points. 

The SM and DDM introduced in Sections \ref{sec:similarity_intro} and \ref{sec:distance_intro} were applied to each pair of movies, respectively. Their results are shown in Table \ref{tab:type1_results}. The results of the combined measure are also shown in Table \ref{tab:type1_results} for comparison purposes, and will be introduced in the next section. For each pair, the FS $A$ was given as the first parameter for the measure, and the FS $B$ was given as the second parameter.
%$f_s$ (\ref{eq:fused_similarity}) & 0.008 & 0.200  & 0.024 & 0.043  & 0.0   & 0.0   & 0.889 \\
      %$f_d$ (\ref{eq:fused_distance})   & 3.151 & -1.211 & 2.709 & -2.486 & 2.183 & 3.167 & 0.006 \\ \bottomrule

\begin{table}[h!]
\setlength{\tabcolsep}{4pt}
\caption{Values given by SMs and DMs on the FSs in Fig. \ref{fig:demonstration_T1_sets}}
  \begin{center}
    \begin{tabular}{  c  c  c  c  c  c  c  c  }
      \toprule
      Fig. \ref{fig:demonstration_T1_sets} - part: 
                                    & a      & b      & c     & d      & e     & f     \\ \midrule
      Similarity (\ref{eq:jaccard}) & 0.050  & 0.067  & 0.170 & 0.242  & 0.0   & 0.892 \\
      Distance (\ref{eq:CR_haus})   & -3.628 & 2.936  & 2.723 & -1.999 & 3.258 & 0.169 \\ 
      Comparative (\ref{eq:fused})  & -0.915 & 0.864  & 0.857 & -0.806 & 0.938 & 0.072 \\ \bottomrule
    \end{tabular}
  \end{center}
  \label{tab:type1_results}
\end{table}
%Fused Comp (\ref{eq:fused})   & -0.085 & 0.136  & 0.143 & -0.194 & 0.062 & 0.928 \\ 

The following is a discussion of the results for similarity (\ref{eq:jaccard}) and distance (\ref{eq:CR_haus}) in Table \ref{tab:type1_results} for the FSs in Fig. \ref{fig:demonstration_T1_sets}. For each case, a brief discussion highlights where both measures contribute information that is particularly helpful when considered together.

\begin{figure}[t!]
  \centering
  \subfigure[]
  {
      \includegraphics[height=4.4cm, width=6cm]{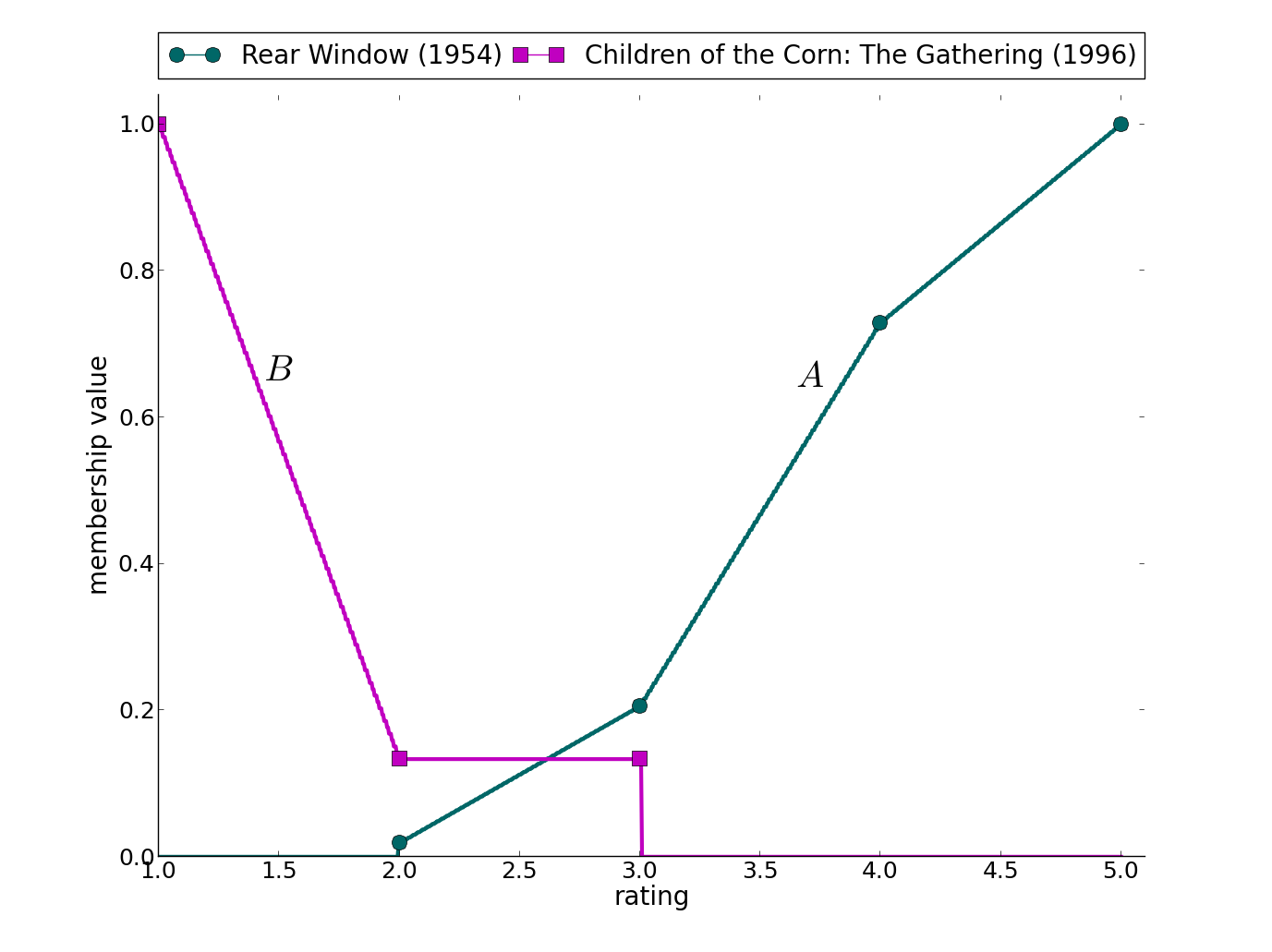}
      \label{fig:demo_a}
  }
  \subfigure[]
  {
      \includegraphics[height=4.4cm, width=6cm]{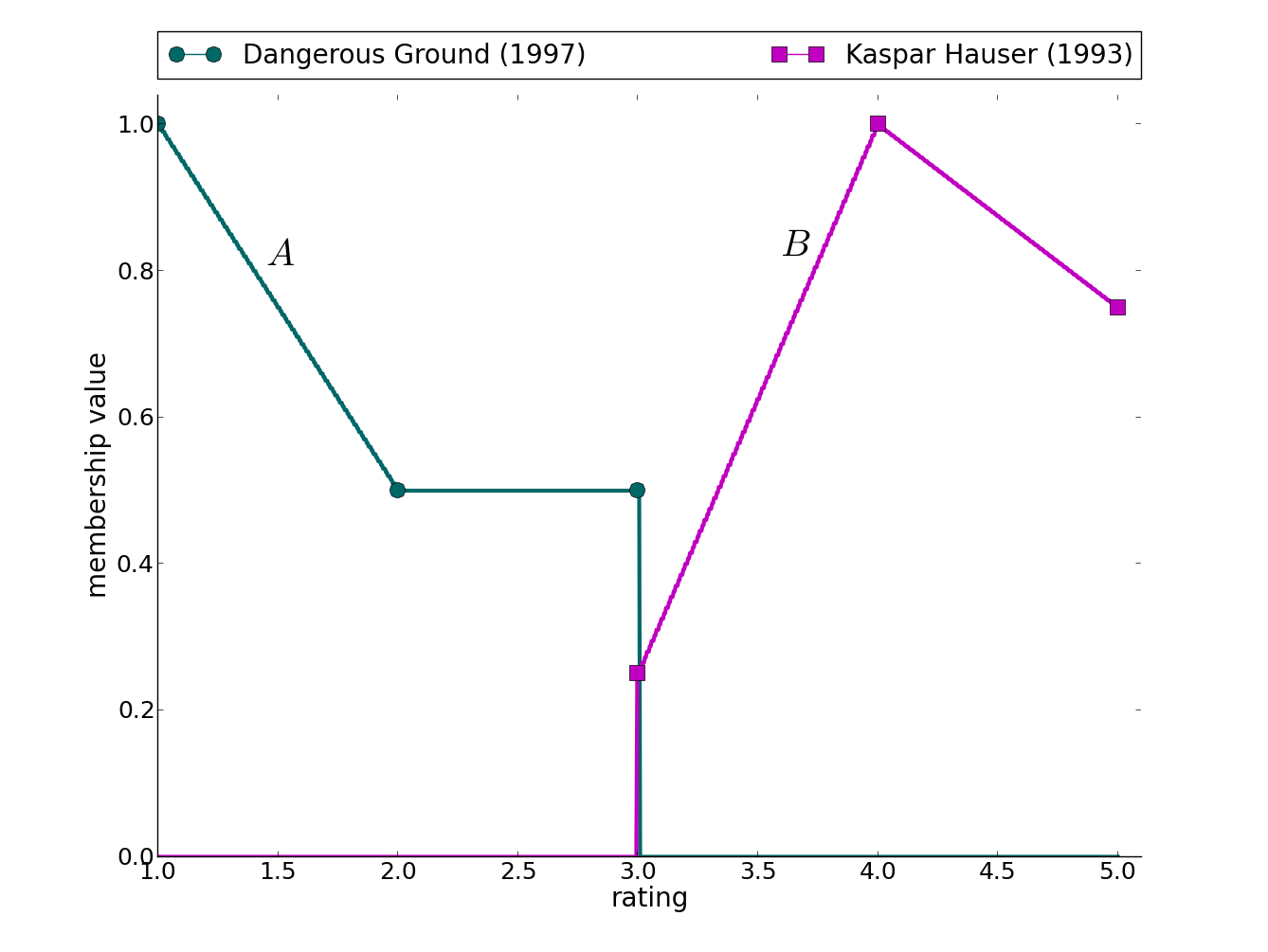}
      \label{fig:demo_b}
  }
  \subfigure[]
  {
      \includegraphics[height=4.4cm, width=6cm]{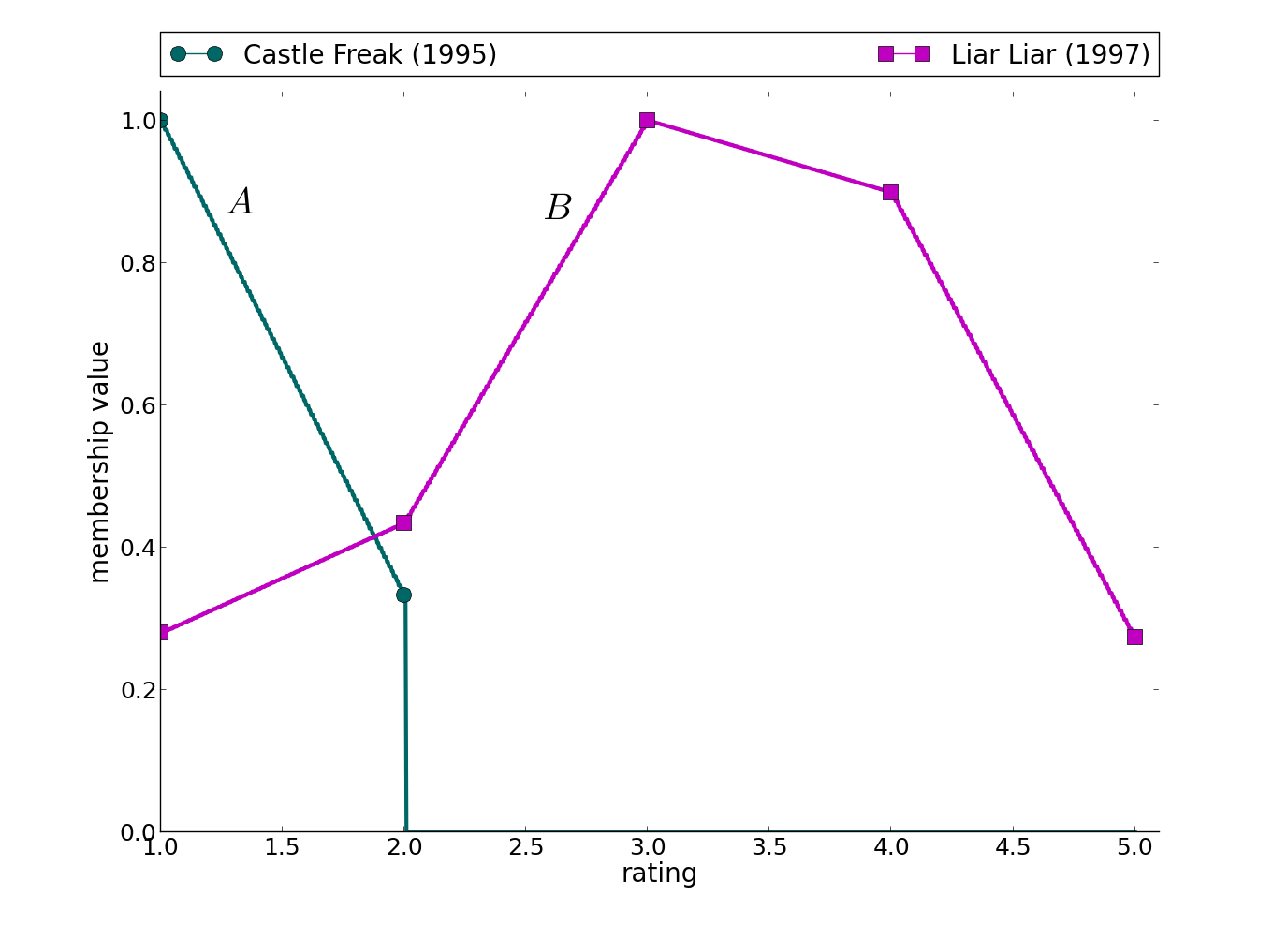}
      \label{fig:demo_c}
  }
  %\caption{Fuzzy sets used to demonstrate the attributes of SMs and DMs in Table \ref{tab:type1_results}.}
  %\label{fig:demonstration_T1_sets}
\end{figure}

\textbf{Sets \textit{a} \& \textit{b}} For the sets in Fig. \ref{fig:demo_a} and \ref{fig:demo_b}, the SM indicates that the FSs are almost disjoint, but there is a small degree of similarity between them. However, there is no indication of where this similarity lies and how much the FSs differ. One can, however, see that the sign of the DM may be helpful to indicate the actual region of similarity. In this case, the direction of the DM tells us that the similarity is likely towards the lower end of the UoD of the FS $A$ for Fig. \ref{fig:demo_a} and the higher end of $A$ for Fig. \ref{fig:demo_b}.

\textbf{Sets \textit{c} \& \textit{d}} The SM indicates a small difference in similarity between the sets of \textit{c} and the sets of \textit{d}, but it tells us very little else; in both cases, there is a small amount of overlap but we do not know where. However, the DM reveals that this overlap is to the right of the FS $A$ for \textit{c}, and to the left $A$ for \textit{d}.

\textbf{Sets \textit{e}} In this case, the SM indicates that there is no similarity between the FSs, i.e. they are disjoint, and the DM indicates that there is a large amount of distance between the FSs.

\textbf{Sets \textit{f}} Both the SM and DM are able to identify when two FSs are identical or, in this case, almost identical. For the results of Fig. \ref{fig:demo_f}, each measure indicates that the membership functions of both FSs are very close to each other.

Given the results above, it is clear that SMs and DMs are each unique functions with distinct properties. This results in the common necessity to choose between both types of measure or indeed to apply both individually. While the application of both measures individually as conducted here can provide some insight, it can be challenging to interpret the two distinct outputs simultaneously for given FSs. 

In the next section, the measures are combined into a single measure resulting in a single value, which can be used to determine the similarity, distance and direction between FSs.

%%%%%%%%%%%%%%%%%%%%%%%%%%%%%%%%%%%%%%%%%%%%%%%%%%%%%%%%%%%%%%%%%%%%%%%%%%%%%%%%%%%%%%%%%%%%%%%

\section{Combining Measures}
\label{sec:combining_measures}
The comparative measure removes the need to choose between measures of similarity or distance, and by combining both measures it creates a more detailed comparison of FSs. Both of these aspects are particularly important in cases where a potentially large number of FSs are generated from data. In such cases, an appropriate decision between the individual measures (and/or joint result interpretation) cannot be conducted by a human expert but has to be done automatically. Thus, a single measure is proposed to provide a detailed comparison of FSs.

\subsection{A Single Comparative Measure}
Note that both measures commonly yield results in different domains; SMs within $[0,1]$ and DMs within $\mathbb{R}$ (or $\mathbb{R}^+$ if it is non-directional). A decision must therefore be made as to which domain will be used for the results of the combined measure. The following presents a measure which yields results in $[0,1]$, for which the value 0 indicates minimum distance/maximum similarity, and the value 1 indicates maximum distance/minimum similarity. 

To fuse the measures, it is important to consider that similarity and distance represent two fundamentally different comparisons of FSs; i.e. both measures measure ``opposite'' concepts. The SM indicates how \textit{similar} or how \textit{close} two FSs are placed, and the DM indicates how \textit{far apart} they are positioned. To fuse these measures they must both represent the same concept, either both \textit{similarity/closeness} or both \textit{dissimilarity/distance}. The following considers the latter case.

As similarity is within the domain $[0, 1]$, to achieve a measure of dissimilarity for the combined measure the complement of the SM may be used (i.e. $1 - s(A,B)$) \cite{Lipkus1999tanimoto}. This can then be used in conjunction with the DM.

Considering the DM is within $\mathbb{R}$, it should be changed such that the result falls within $[0,1]$ to enable a meaningful fusion of both measures. To alter the result, it is necessary to take into account the UoD in which the measure has been applied. For example, if the UoD is in $\{1, 2, 3, 4, 5\}$ then the maximum distance that can be achieved is 4. The result of the DM may be given as a ratio to the maximum possible distance. Taking the above into account, $\frac{d(A,B)}{\lambda}$ is used to obtain a ratio of closeness from the DM, where $\lambda$ is the largest possible distance within the UoD. For a finite UoD $X$ described as $\{x_1, x_2, ...., x_n\}$, $\lambda$ will be $x_n - x_1$. 

Given the above, the OWA provides a reasonable approach to fusing both measures ((\ref{eq:jaccard}) and (\ref{eq:CR_haus})). (\ref{eq:fused}) presents the comparative measure as an OWA based aggregation of the measures of similarity and distance for FSs $A$ and $B$:

\begin{equation}
c(A,B) =
 \begin{cases}
  F{\scriptstyle \big( (1-s(A,B))}, \big( \frac{d(A,B)}{\lambda} \big)\big), & d(A,B) \geq 0 \\
  F{\scriptstyle \big( -(1-s(A,B))}, \big( \frac{d(A,B)}{\lambda} \big)\big), & \text{otherwise}
\end{cases}
 \label{eq:fused}
\end{equation}
where $F$ is an OWA as shown in (\ref{eq:OWA}) with weights $w=\{0.7, 0.3\}$ and $d$ is the DDM (\ref{eq:CR_haus}) with (\ref{eq:interval_haus_with_sign}). 

The weights $w=\{0.7, 0.3\}$ are chosen such that the largest of the dissimilarity measure and normalised DM within (\ref{eq:fused}) is assigned the weight 0.7, and the smallest is assigned the weight 0.3. Note that the absolute values of the measures are used when assigning the weights, thus a measure of -0.45 is considered larger than a measure of 0.3. These weights have been determined heuristically as outlined in Section \ref{sec:owa_weights}, and in the future other ways of determining such weights may be investigated. 

Note that if the result from the DDM gives a negative value then the result of (\ref{eq:fused}) will also be a negative value. Likewise, if the DDM gives a positive value then the result of the combined measure will also be positive.

In (\ref{eq:fused}), a value of 0 represents identical FSs, as proven in theorem 1 below, and a value of 1 (or -1) represents the maximum distance possible of two disjoint FSs. If one wishes to have the value 1 to represent identical FSs, the complement of (\ref{eq:fused}) may be used as
\begin{equation}
  c'(A,B) = 
  \begin{cases}
    1 - c(A,B),  & \text{if } c(A,B) \geq 0 \\
    -1 - c(A,B), & \text{otherwise}
  \end{cases}
  \label{eq:fused_comp}
\end{equation}
Note that (\ref{eq:fused_comp}) maintains the direction according to the comparative measure (\ref{eq:fused}).

Within (\ref{eq:fused}), the measure of similarity is altered such that it reflects dissimilarity or distance. Note that this is just one method of combining the measures proposed because of both its simplicity and its ability to represent both similarity and distance as demonstrated in the examples within the next section. Another method, for example, is the special case where the weights are both 0.5, resulting in the standard average of both measures. It is also possible that the result may be altered to yield results in the domain $\mathbb{R}$ by multiplying the SM by the value $\lambda$ and fusing the result with the unaltered DM.

\subsection{Choosing the OWA Weights}
\label{sec:owa_weights}
\begin{figure}[t]
  \centering
  \includegraphics[height=4.7cm, width=6cm]{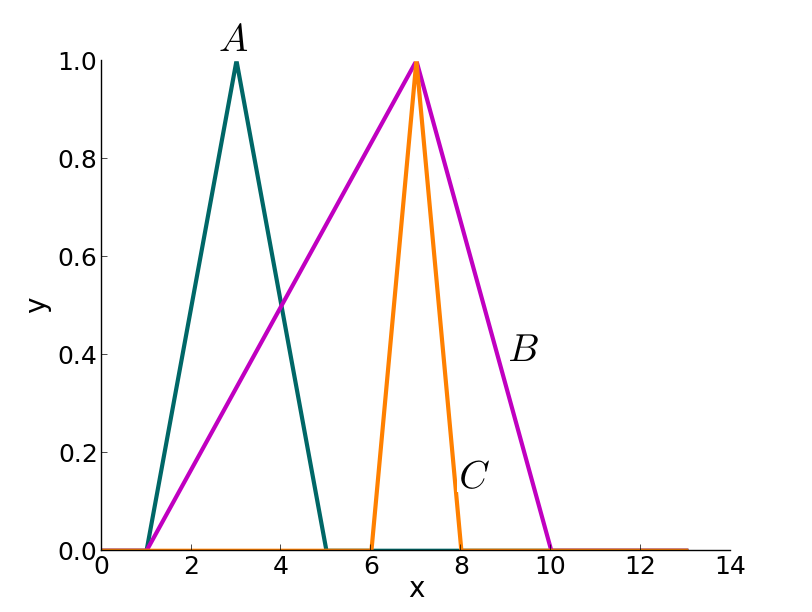}
  \caption{Three FSs, $A$, $B$ and $C$.}
  \label{fig:weights_example}
\end{figure}

The following discusses how the weights of the OWA operator (\ref{eq:OWA}) may be chosen for the comparative measure (\ref{eq:fused}). Referring to Fig. \ref{fig:weights_example}, the FS pairs $(A, B)$ and $(A, C)$ are compared. According to the DM (\ref{eq:CR_haus}) the distance for both pairs is 0.331. However, according to the SM (\ref{eq:jaccard}) the similarity of $(A, B)$ is 0.182, and the similarity of $(A, C)$ is 0.0. Due to the DM giving the same result for $(A, B)$ and $(A, C)$ one would assume that the $B$ and $C$ are the same FS. It is only by also referring to the SM (or by viewing the FSs) that it becomes clear that the FSs are different. By using the comparative measure, however, it is possible to distinguish between different pairs of FSs which give equal values of similarity or distance. It is also important to note that, by using the comparative measure, this can be confirmed by using a single measure; a user does not have to check the results of both the SM and DM to ensure pairs of FSs are different. 

The weights of the comparative measure play an important role in determining the difference between different pairs of FSs which result in equal values from a measure. Table \ref{tab:weights_example} shows the difference between pairs $(A,B)$ and $(A,C)$ using a variety of weights. As the first weight increase in value the difference between the two pairs also increases; the results begin to signify that $B$ is closer to $A$ than $C$ is to $A$. This is because the dissimilarity measure gives 1 for disjoint sets (such as pairs $(A,C)$) and thus, in such cases, will always be given the first weight. As the first weight increases in value the overall measure is, in effect, placing more importance on the fact that the sets are disjoint.

However, it is unhelpful to have too large of a value for the first weight. If the first weight equals 1 then the output of the combined measure will always equal 1 for disjoint sets. Considering this, the first weight should be low enough such that it is possible to distinguish between different pairs of disjoint sets. However, it must also be large enough such that it is possible to make a distinction between FSs which give equivalent values of similarity or distance, such as those in Fig. \ref{fig:weights_example}.

Ideally, when the FSs are known beforehand, the weights should be tuned such that the widest range of values are given by the measure. This decreases possible confusions over pairs of FSs which would give close or identical values from a single measure. However, if the weights cannot be tuned, the weights $\{0.7, 0.3\}$ are ideal and are used throughout this paper. This is because tests showed that these weights are useful for preventing disjoint FSs from resulting in a lower distance/dissimilarity than non-disjoint FSs.

\begin{table}
\caption{Comparative measure on the FSs within Fig. \ref{fig:weights_example} using different weights (* indicates the chosen weights within this paper).}
  \begin{center}
    \begin{tabular}{  c  c  c  c  }
      \toprule
      \ Weight 0 & Weight 1 & c(A,B) & c(A,C) \\ \midrule
      \ 0.0 & 1.0 & 0.331 & 0.331 \\
      \ 0.1 & 0.9 & 0.380 & 0.398 \\
      \ 0.2 & 0.8 & 0.428 & 0.465 \\
      \ 0.3 & 0.7 & 0.477 & 0.532 \\
      \ 0.4 & 0.6 & 0.526 & 0.598 \\
      \ 0.5 & 0.5 & 0.574 & 0.665 \\
      \ 0.6 & 0.4 & 0.623 & 0.732 \\
      *0.7 & 0.3 & 0.672 & 0.799 \\
      \ 0.8 & 0.2 & 0.721 & 0.866 \\
      \ 0.9 & 0.1 & 0.769 & 0.933 \\
      \ 1.0 & 0.0 & 0.818 & 1.0 \\ \bottomrule
    \end{tabular}
  \end{center}
  \label{tab:weights_example}
\end{table}

\subsection{Properties of the combined measure}
This Section introduces and proves the properties of the combined measure (\ref{eq:fused}).

\begin{thm}[Self-identity]
The comparative measure (\ref{eq:fused}) follows the property of self-identity. That is, $c(A,B) = 0 \Longleftrightarrow A = B$.
\end{thm}
\begin{IEEEproof}
 If $A = B$ then $s(A,B)=1$ according the property of reflexivity, and so $w(1-s(A,B))=0$. \\
 Also, if $A=B$ then $d(A,B) = 0$ according to the property of self-identity, and so $w \bigg( \frac{d(A,B)}{\lambda} \bigg) = 0$ for any $w$. Thus $c(A,B) = 0$ if $A=B$.\\
 Alternatively, if $A \neq B$ then $s(A,B) \neq 1$ and $d(A,B) \neq 0$, thus $c(A,B) \neq 0$.
\end{IEEEproof}

\begin{thm}[Symmetry]
The comparative measure (\ref{eq:fused}) follows the property of symmetry. That is, $c(A,B) = c(B,A)$.
\end{thm}
\begin{IEEEproof}
 If the SM and DM that are aggregated are both symmetrical, then the same values will be given to the comparative measure for both $c(A,B)$ and $c(B,A)$, thus the comparative measure is also symmetrical.
\end{IEEEproof}

\begin{thm}[Partial Symmetry]
Where the DDM (\ref{eq:interval_haus_with_sign}) is used with the comparative measure (\ref{eq:fused}), the property of partial symmetry holds. That is $|c(A,B)| = |c(B,A)|$, and $c(A,B) \neq c(B,A)$ where $A \neq B$
\end{thm}
\begin{IEEEproof}
When the distance is a positive value, the values of both distance and dissimilarity given to the OWA are in the positive domain. Likewise, where the distance is negative, both inputs given to the OWA are in the negative domain. In each case, the absolute values of the positive and negative inputs are the same, thus the absolute values of the outputs are also the same, and the sign of the final value is in same domain as the input values.
\end{IEEEproof}

\begin{thm}[Separability]
The result of the comparative measure is always greater than or equal to zero, i.e. $c(A,B) \geq 0$ when aggregating with the non-DDM.
\end{thm}
\begin{IEEEproof}
 Given that $1-s(A,B) \in [0,1]$, and $\frac{d(A,B)}{\lambda} \in [0,1]$, as $\lambda$ never exceeds the maximum value of $d(A,B)$, it follows that $c(A,B) \in [0,1]$ thus $c(A,B) \geq 0$.\\
\end{IEEEproof}

Note, however, if the DDM is used to construct the comparative measure, then $c(A,B) \in [-1, 1]$. Thus, separability does not apply where the DDM is used.

\begin{thm}[Transitivity]
The comparative measure (\ref{eq:fused}) follows the property of transitivity. That is, if $A \subseteq B \subseteq C$, then $d(A, B) \leq d(A, C)$
\end{thm}
\begin{IEEEproof}
 Given that both the dissimilarity measure and the DM follow transitivity, when they are aggregated the resulting comparative measure also follows transitivity.
\end{IEEEproof}

\begin{thm}[Triangle Inequality]
The comparative measure (\ref{eq:fused}) follows the property of triangle inequality. That is, $ c(A, C) \leq c(A, B) + c(B, C) $ 
\end{thm}
\begin{IEEEproof}
 Given that both the dissimilarity measure \cite{Lipkus1999tanimoto} and the DM follow triangle inequality, when they are aggregated the resulting comparative measure also follows triangle inequality.
\end{IEEEproof}

The complement of the comparative measure (\ref{eq:fused_comp}), likewise, follows the properties of symmetry, separately, transitivity and triangle inequality. However, the complement does not satisfy self-identity and instead follows reflexivity ($c'(A,B)=1 \Leftrightarrow A=B$). This is because the complement uses 1 to indicate identical FSs, where as the comparative measure uses 0 instead. It is trivial to see from theorem 1 that the complement of the comparative measure satisfies reflexivity.

Note that the comparative measure does not follow the property of overlapping (i.e. if $A \cap B \neq \emptyset$, then $c(A, B) > 0$, otherwise $c(A, B) = 0$) unless the weights $w=[1.0, 0]$ are given, such that the maximum weight is given to the dissimilarity measure when the FSs are identified as disjoint.

\section{Demonstrations}
\label{sec:demonstrations}
Examples of the comparative measure (\ref{eq:fused}) are given in Table \ref{tab:type1_results} in which the measure is applied to the FSs in Fig. \ref{fig:demonstration_T1_sets}. A demonstration and discussion of the comparative measure in an applied context are presented next, and compared against using a single measure of similarity or distance.

\subsection{Demonstration - MovieLens}
The following is a discussion of comparisons between the FSs in Fig. \ref{fig:demonstration_T1_sets} according to the comparative measure (\ref{eq:fused}), the results of which are shown in Table \ref{tab:type1_results}.

\textbf{Sets \textit{a} \& \textit{b}} The results of \textit{a} and \textit{b} in Table \ref{tab:type1_results} have a high degree of dissimilarity/distance. Additionally, the sign of the comparative measure shows in which direction the dissimilar regions of the FSs reside. To the left of $A$ in the case of the FSs in \textit{a} and to the right of $A$ in the case of the sets in \textit{b}. The comparative measure also shows that the FSs within \textit{b}  are closer than those in \textit{a}. 

\textbf{Sets \textit{c} \& \textit{d}} The FSs within \textit{c} and \textit{d} both have a slightly increased degree of dissimilarity/distance than indicated by the original SM and DM. This is due to a large difference between the range of elements contained within the sets, which increases the dissimilarity according to the measure. In both cases, the FS $B$ covers the range $[1,5]$ whereas $A$ only covers $[1,2]$ in \textit{c}, and $[4,5]$ in \textit{d}. It can also now be observed, by using the comparative measure alone, that $B$ is to the right of $A$ in \textit{c}, and is to the left of $A$ in \textit{d}. The ordering and direction of the distance between \textit{c} and \textit{d} are the same using both the DM and the comparative measure.

\textbf{Sets \textit{e}} The FSs within \textit{e} are disjoint and were thus given the value 0 by the Jaccard SM. However, the comparative measure gives a non-zero value for \textit{e}. Note that this value is still the largest dissimilarity/distance compared to the other pairs within Table \ref{tab:type1_results}. This value also helps to identify the direction of the FSs indicating that the FS $B$ is to the right of $A$.

\textbf{Sets \textit{f}} The comparative measure indicates with a high degree of certainty that the FSs of \textit{f} are nearly identical.

A possible application of the comparative measure is to the problem of ranking. Comparing the comparative measure against the DM, which may also be used for ranking \cite{Cheng1998307}, the ordering of the FSs differs. By observing the absolute values of the measures, according to the DM the most distant pair is \textit{a} and the second most distant is \textit{e}, however, it is the other way round according to the comparative measure. This is because the SM indicates there is some similarity between the FSs within Fig. \ref{fig:demo_a}, which causes the comparative measure to decrease in dissimilarity/distance. However, the FSs in Fig. \ref{fig:demo_e} are disjoint, so the dissimilarity/distance remains high. This leaves the FSs with no similarity as the most distant. One could argue that this is an expected result of the comparative measure because the FSs within \textit{e} are disjoint where as the FSs within \textit{a} are not disjoint. Thus the measure may be considered more intuitive as it is natural to consider the sets in \textit{e} as being more distant than the sets in \textit{a}.

As stated earlier, the unique properties of similarity and distance enable the measures to be applied to a wide variety of fields, and the same can be said for the comparative measure, which, as demonstrated, can be used in terms of a measure of similarity and a measure of distance. For example, with the FSs in Fig. \ref{fig:demonstration_T1_sets} the comparative measure may be used to find similarly rated films by choosing FSs with a low value of dissimilarity/distance, or it may be used to rank the film ratings by ordering the results of the measure. 

\subsection{Demonstration - Classification}
\label{sec:demo_classification}
\begin{figure}[t]
  \centering
  \includegraphics[height=4.5cm, width=6cm]{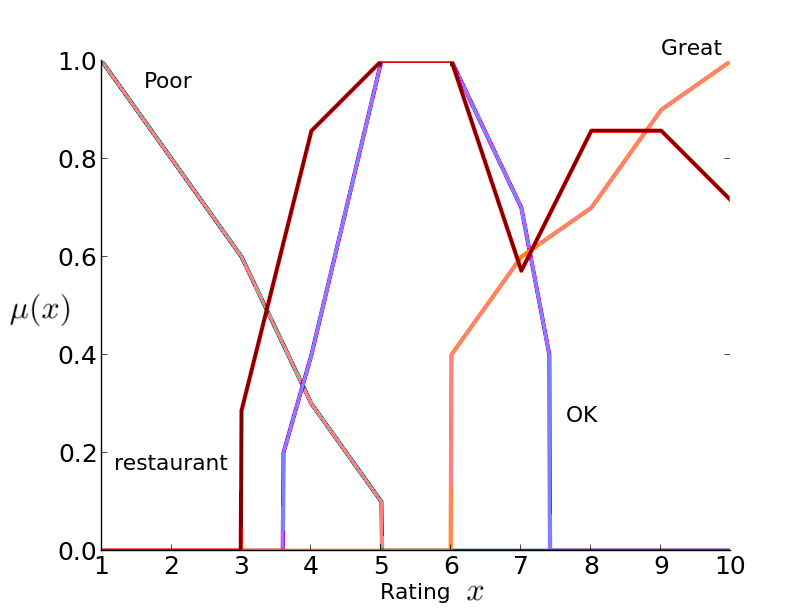}
  \caption{Three FSs modelling degrees of ambience, with a FS representing the ambience of a given restaurant.}
  \label{fig:restaurant_example}
\end{figure}

This section presents a synthetic example of the comparative measure applied to the problem of classification. In this example, three initial FSs are given which represent three different descriptions. In this case, they each represent different levels of ambience within an establishment on a scale from 1 to 10. These levels, as shown in Fig. \ref{fig:restaurant_example}, are labelled as \textit{Poor}, \textit{OK} and \textit{Great}. Given a FS representing the ambience of a restaurant, as shown in Fig. \ref{fig:restaurant_example}, the aim is to classify which description best fits the restaurant.

In Table \ref{tab:restaurant_example}, comparisons are given between the ambience of the restaurant and the descriptions. The measures of similarity (\ref{eq:jaccard}) and distance (\ref{eq:CR_haus}) are shown, as well as the complement of the comparative measure (\ref{eq:fused_comp}). For each measure, the word model is given as the first parameter, and the restaurant is given as the second parameter. The complement of the comparative measure is given to match the SM, such that both measures give the value 1 for identical FSs.

\begin{table}[t]
\caption{Similarity and distance between the restaurant and word models in Fig. \ref{fig:restaurant_example}}
  \begin{center}
    \begin{tabular}{  c  c  c  c  }
      \toprule
                                         & Poor   & OK     & Great \\ \midrule
      Similarity (\ref{eq:jaccard})      & 0.081 & 0.493 & 0.469  \\
      Distance (\ref{eq:CR_haus})        & 5.573 & 1.064 & -3.360 \\
      Comparative (\ref{eq:fused_comp}) & 0.171 & 0.609 & -0.516 \\ \bottomrule
    \end{tabular}
  \end{center}
  \label{tab:restaurant_example}
\end{table}

According to the SM (\ref{eq:jaccard}), the restaurant's ambience is similar to the descriptions \textit{poor}, \textit{OK} and \textit{great} to the degree 0.081, 0.493 and 0.469, respectively. It is clear from these results that the restaurant's ambience cannot be described as \textit{poor}, however, it is almost equally valid that it may be described as \textit{OK} or \textit{Great}. 

The DM (\ref{eq:CR_haus}), however, gives a clearer view of which FS the restaurant most closely matches; the restaurant has a smaller distance to \textit{OK} than to \textit{Great}. Thus, by fusing the distance and similarity as in (\ref{eq:fused}) and (\ref{eq:fused_comp}), a more distinct match is achieved. Now, it is clear from the results of the comparative measure in Table \ref{tab:restaurant_example} that the restaurant most closely matches \textit{OK} ambience to the degree of 0.609, whereas it only matches \textit{Great} by 0.516 and \textit{Poor} by 0.171. It can now be determined with greater certainty that \textit{OK} is the correct classification. It should be noted that this can be determined by observing a single measure (\ref{eq:fused_comp}), rather than viewing the SM and DM separately.

\section{Conclusions}
\label{sec:conclusions}
\enlargethispage{-0.5cm}
This paper has introduced a novel measure, referred to as a comparative measure, which analyses and compares FSs by combining a SM and DM. When these measures are viewed separately the results may be difficult and time-consuming to interpret as similarity and distance each measure fundamentally different concepts. By joining the measures together, the comparison of FSs is simplified by reducing any ambiguity in the results. Additionally, compared to a single measure, the combined measure provides is a richer comparison as it may be swayed towards a preference in representing similarity or distance. This is especially useful for the automatic comparison of a large number of FSs which have been constructed from data. Additionally, through using an OWA operator, it is possible to refine the weights to further alleviate ambiguous values resulting from the original measures.

Demonstrations using data-driven FSs have shown that the comparative measure may be applied in terms of both similarity and distance, and as such may be applied to applications of these measures. Though the demonstrations have been applied to type-1 FSs only, as the comparative measure uses the outputs of the SM and DM, it may also be applied to type-2 FSs, where the original measures are a type-2 SM and DM.

Future work will look at measures which indicate similarity and distance as a FS, which better reflects the uncertainty inherent in FSs.

\bibliographystyle{IEEEtran}
\bibliography{IEEEabrv.bib,papers.bib}

\end{document}